\documentclass[10pt,twocolumn,letterpaper]{article}

\usepackage{cvpr}        
\usepackage[accsupp]{axessibility}  
\usepackage{graphicx}
\usepackage{amsmath}
\usepackage{amssymb}
\usepackage{booktabs}
\usepackage[dvipsnames,table,xcdraw]{xcolor}
\usepackage{color}

 \usepackage{multirow}
\usepackage[pagebackref,breaklinks,colorlinks]{hyperref}

\usepackage{algorithm}
\usepackage{algorithmic}

\usepackage[capitalize]{cleveref}
\crefname{section}{Sec.}{Secs.}
\Crefname{section}{Section}{Sections}
\Crefname{table}{Table}{Tables}
\crefname{table}{Tab.}{Tabs.}

\def\cvprPaperID{*****} 
\def\confName{CVPR}
\def\confYear{2023}

\begin{document}

\title{t-RAIN: Robust generalization under weather-aliasing label shift attacks}

\author{Aboli Marathe$^1$, Sanjana Prabhu$^2$\\
$^1$Machine Learning Department,\\
$^2$Department of Electrical and Computer Engineering,\\
Carnegie Mellon University\\
Pittsburgh, PA 15213\\
{\tt\small abolim@cs.cmu.edu,sprabhu2@andrew.cmu.edu}
}
\maketitle

\begin{abstract}

In the classical supervised learning settings, classifiers are fit with the assumption of balanced label distributions and produce remarkable results on the same. In the real world, however, these assumptions often bend and in turn adversely impact model performance. Identifying bad learners in skewed target distributions is even more challenging. Thus achieving model robustness under these "label shift" settings is an important task in autonomous perception. In this paper, we analyze the impact of label shift on the task of multi-weather classification for autonomous vehicles. We use this information as a prior to better assess pedestrian detection in adverse weather. We model the classification performance as an indicator of robustness under 4 label shift scenarios and study the behavior of multiple classes of models. We propose t-RAIN a similarity mapping technique for synthetic data augmentation using large scale generative models and evaluate the performance on DAWN dataset. This mapping boosts model test accuracy by 2.1, 4.4, 1.9, 2.7 \% in no-shift, fog, snow, dust shifts respectively. We present state-of-the-art pedestrian detection results on real and synthetic weather domains with best performing 82.69 AP (snow) and 62.31 AP (fog) respectively. 
\end{abstract}

\section{Introduction}

\begin{figure}[h!]
    \centering
    \includegraphics[width=0.46\textwidth,height=9cm]{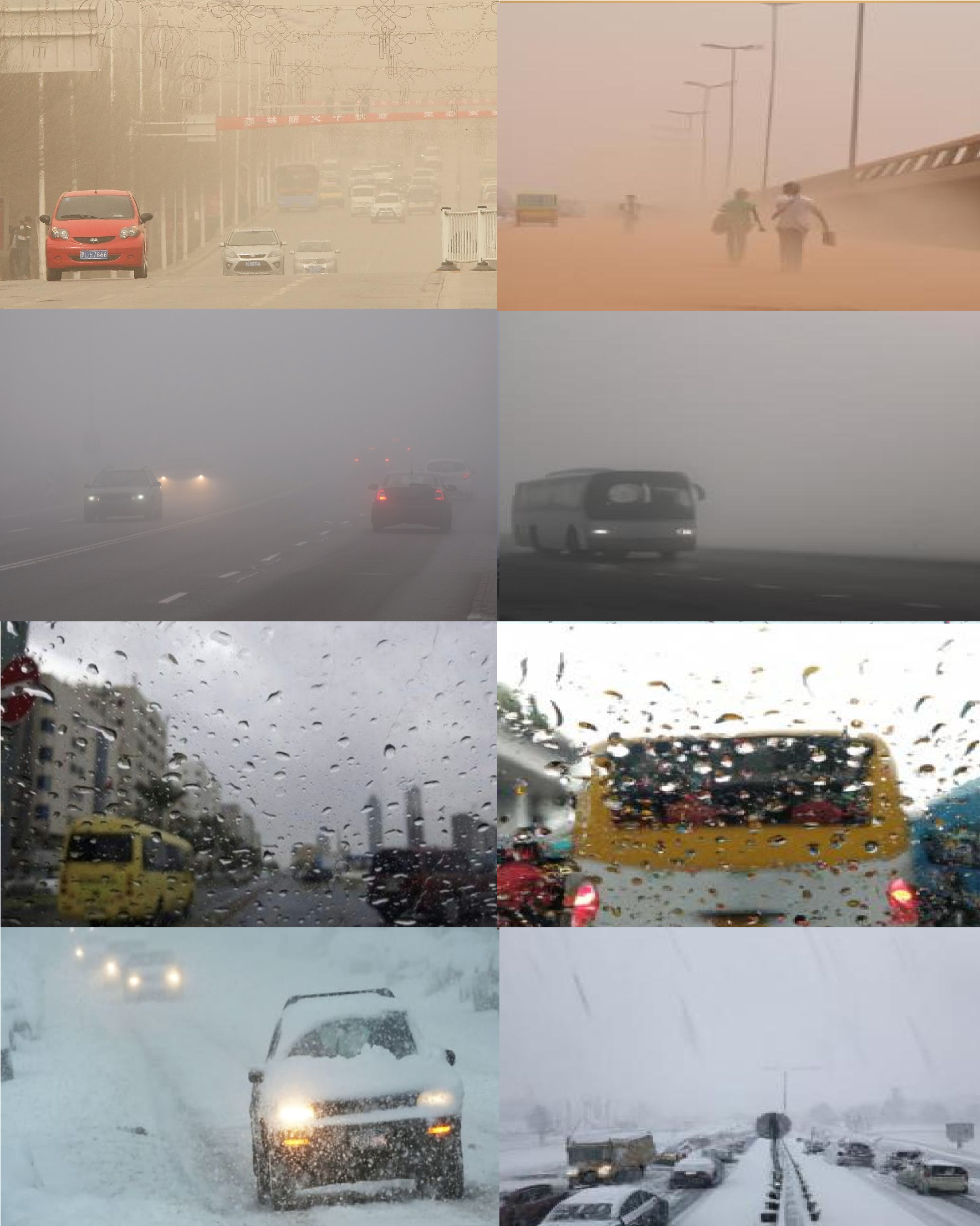}
    \caption{\textbf{Sim2Real Detection:} DAWN-WEDGE (Real-Synthetic) Data Samples Depicting Adversarial Weather Conditions Including Dust (Tornado, Sandstorms), Fog (Mist, Haze, Fog), Rain and Snow in Autonomous Driving Scenes.}
    \label{fig1}
\end{figure}

\begin{table*}[]
\begin{center}

\resizebox{1.9\columnwidth}{!}{\begin{tabular}{lllllllllllllll}
\hline
\multicolumn{1}{l|}{} & \multicolumn{8}{l|}{\textbf{Real Data (DAWN Dataset)}} & \multicolumn{6}{l}{\textbf{Synthetic Data (WEDGE Dataset)}} \\ \cline{2-15} 
\multicolumn{1}{l|}{\multirow{-2}{*}{\textbf{Model}}} & \textbf{car} & \textbf{person} & \textbf{bus} & \textbf{truck} & \textbf{T-4 AP} & \textbf{mc} & \textbf{bicycle} & \multicolumn{1}{l|}{\textbf{mAP}} & \textbf{car} & \textbf{person} & \textbf{bus} & \textbf{truck} & \textbf{van} & \textbf{mAP} \\ \hline
\multicolumn{15}{l}{\textbf{Prior Art}} \\ \hline
\multicolumn{1}{l|}{Multi-weather city \cite{ms}} & - & - & - & - & 21.20 (39.19) & - & - & \multicolumn{1}{l|}{(39.19)} & - & - & - & - & - & - \\
\multicolumn{1}{l|}{RoHL \cite{saikia2021improving}} & - & - & - & - & - & - & - & \multicolumn{1}{l|}{28.80} & - & - & - & - & - & - \\
\multicolumn{1}{l|}{Transfer Learning \cite{marathe2022rain}} & 7.00 & 8.00 & 7.00 & - & 5.50 & - & 0.00 &  \multicolumn{1}{l|}{-} & - & - & - & - & - & - \\
\multicolumn{1}{l|}{Data Augmentation \cite{marathe2022rain}} & 6.00 & 4.00 & 3.00 & 0.00 & 26.25 & - & \textbf{92.00} & \multicolumn{1}{l|}{-} & - & - & - & - & - & - \\
\multicolumn{1}{l|}{\begin{tabular}[c]{@{}l@{}}Weather-\\ Night GAN \cite{marathe2022restorex}\end{tabular}} & 48.00 & 0.00 & 0.00 & 0.00 & 12.00 & - & - & \multicolumn{1}{l|}{-} & - & - & - & - & - & - \\
\multicolumn{1}{l|}{Ensemble Detectors \cite{a3}} & 52.56 & 52.34 & 21.73 & 13.71 & 35.08 & 35.51 & 23.29 & \multicolumn{1}{l|}{32.75} & - & - & - & - & - & - \\ \hline \hline
\multicolumn{15}{l}{\textbf{Evaluation on DAWN-All}} \\ \hline
\multicolumn{15}{l}{\textbf{Trained on Good Weather Data (COCO \cite{lin2014microsoft})}} \\ \hline

\multicolumn{1}{l|}{\begin{tabular}[c]{@{}l@{}}FasterRCNN   \\ MobileNet \\ Large 320 \cite{ren2015faster,howard2017mobilenets}\end{tabular}} &37.56	&	34.93	&	20.90	&	12.91	&	26.57	&	23.15	&	18.95	&	\multicolumn{1}{l|}{24.73} & 34.10	&	36.26	&	39.35	&	16.05	&	0.00	&	25.15\\
\multicolumn{1}{l|}{\begin{tabular}[c]{@{}l@{}}FasterRCNN   \\ MobileNet\\ Large \cite{ren2015faster,howard2017mobilenets}\end{tabular}} & 60.64	&	55.96	&	32.78	&	23.66	&	43.26	&	38.55	&	28.75	&	\multicolumn{1}{l|}{40.05} & 35.34	&	39.52	&	35.83	&	25.43	&	0.00	&	27.22\\
\multicolumn{1}{l|}{FasterRCNN ResNet 50\cite{ren2015faster}} &  \textbf{69.13}	&	\textbf{70.31}	&	\textbf{38.64}	&	30.54	&	\textbf{52.15}	&	\textbf{52.17}	&	\textbf{30.56}	&	\multicolumn{1}{l|}{\textbf{48.55}}  & 31.41	&	33.54	&	30.19	&	18.75	&	0.00	&	22.78 \\ \hline

\multicolumn{15}{l}{\textbf{Fine-Tuning on WEDGE}} \\ \hline
\multicolumn{1}{l|}{\begin{tabular}[c]{@{}l@{}}FasterRCNN   \\ MobileNet \\ Large 320 \cite{ren2015faster,howard2017mobilenets}\end{tabular}} & \textbf{39.52}	&	23.97	&	7.81	&	\textbf{22.08}	&	23.34	&	0.00	&	0.00	&	\multicolumn{1}{l|}{15.56} & 40.40	&	43.01	&	49.88	&	31.41	&	10.19	&	34.98 \\
\multicolumn{1}{l|}{\begin{tabular}[c]{@{}l@{}}FasterRCNN   \\ MobileNet \\ Large \cite{ren2015faster,howard2017mobilenets}\end{tabular}} & 59.81	&	34.61	&	14.06	&	\textbf{30.67}	&	34.78	&	0.00	&	0.00	&	\multicolumn{1}{l|}{23.19}  & 52.52	&	\textbf{54.79}	&	\textbf{51.23}	&	50.01	&	7.95	&	43.30\\ 
\multicolumn{1}{l|}{FasterRCNN ResNet 50\cite{ren2015faster}} &  68.09	&	54.29	&	27.48	&	\textbf{35.02}	&	46.22	&	0.00	&	0.00	&	\multicolumn{1}{l|}{30.81} & \textbf{57.48}	&	54.71	&	46.92	&	\textbf{57.43}	&	\textbf{10.49}	&	\textbf{45.41}\\ \hline
\end{tabular}}

\caption{\textbf{Object Detection Benchmarks on DAWN and WEDGE datasets.} The latest work \cite{aa} in this domain presents state-of-the-art benchmark (models trained on Good-weather data and fine-tuned on WEDGE) on DAWN and WEDGE datasets. The pedestrian detection benchmark is \textbf{70.31 AP}  on DAWN and \textbf{54.79 AP} on WEDGE. (54.29 and 54.71 is best WEDGE fine-tuned model benchmark) }
 \label{tab3}
     
\end{center}
\end{table*}

Autonomous perception is notoriously vulnerable to out-of-distribution settings like adverse weather and imagery corruptions. As data from sensors is both limited and often corrupted by natural phenomena, for practical purposes, in-built model robustness is essential for efficient computation. Given the dynamic surroundings and terrains present in everyday driving scenes, building robustness to out-of-distributions settings is an essential feature for vehicular safety and trust. However, modern classifiers are mostly trained on good-weather data due to the abundance and ease of classification, making them vulnerable to adversarial weather attacks like sand, dust, mist, snow, droplets, fog and rain. 

In this work, we treat multi-weather robustness as a supervised learning problem in the standard settings and optimize for best performance. Then we perturb the target distribution to simulate label shift and test this robustness.  The main goal is pedestrian detection under adversarial weather conditions and study of the underlying performance shifts.  Our main contributions include:
\begin{enumerate}
    \item \textbf{Benchmark.} Multi-weather classification benchmark on DAWN dataset. Analysis of model behaviour under limited settings.
    \item \textbf{Label Shift. }Simulation of label shift settings for multi-weather classification. Proposal of t-RAIN algorithm for synthetic data augmentation using VLM prompting.
    \item \textbf{Pedestrian Detection.} We conduct experiments to link the multi-weather classification behaviour by considering the task of pedestrian detection in synthetic and real settings.

\end{enumerate}

\section{Background}

\subsection{Label Shift}

Tackling image corruptions for improved perception has been a long-standing challenge in the field of computer vision \cite{buades2005review,buades2005non,motwani2004survey}. As newer datasets have introduced weather-based corruptions \cite{caesar2020nuscenes,kenk2020dawn,wilson2023argoverse} for improving robustness, the awareness of this subject is on the rise.  Recently, multi-weather robustness has been the focus of several works which proposed ideas like stacking \cite{ms}, ensembles \cite{a3} and image restoration \cite{marathe2022rain}, the performance of classic benchmark models still fail on extreme weather conditions. 
In the history of label shift methods, several works study correcting label shift and generalization in general, especially in unsupervised settings \cite{w1,w2,w3}. The progress in this field is rapid due to the parallel development of autonomous vehicles and need for explainability for trust-worthy AI systems for the future.

\subsection{Adversarial Weather Robustness}

The DAWN dataset \cite{kenk2020dawn}  and WEDGE dataset \cite{aa} present interesting adversarial weather conditions including fog, rain, snow , dust as visible in Figure \ref{fig1}. The most recent benchmark on these datasets \cite{marathe2022rain,aa} presents state-of-the-art results and demonstrates effectiveness of using synthetic data augmentation in the task of overall object detection. 

\subsection{Sim2Real Gap}

The recent development in large vision-language models \cite{ramesh2021zero,ramesh2022hierarchical} and refined generative techniques, have led to creation of more realistic synthetic images. The natural advancement would be adoption of synthetic images to augment limited real-world datasets. However, this adoption is limited by the cost, realism, availability and usability of synthetic images. By incorporating synthetic images \cite{aa} in this work, we demonstrate one positive use case of such images. 

\subsection{Pedestrian Detection}

Finding people in imagery has been a long-standing challenge in computer vision, with several decades of prior work \cite{dalal2005histograms,viola2005detecting,zhang2015filtered} setting up the foundation for examination of finer problems in modern computer vision.  The creation of high-quality datasets \cite{dalal2005histograms,wojek2009multi,ess2008mobile,geiger2012we} was a contributing factor to rapid development of powerful algorithms capable of detection in challenging conditions.  Classical algorithms combined with novel architectures for robust detection were the focus of many works in vision encompassing multi-scale detection, occlusion invariance and  cascaded rejection
classifiers \cite{walambe2021multiscale, walambe2021lightweight,  marathe2021evaluating, cai2016unified,mao2017can,sermanet2013pedestrian,yang2016exploit}.More recently, tracking and detection of pedestrians in the real-world has been solved using a variety of deep networks, algorithmic strategies and forecasting approaches \cite{khurana2021detecting,sigal2004tracking,dendorfer2019cvpr19,dendorfer2020mot20}.

\begin{table*}[]
\begin{tabular}{lllllllllll} \hline
\textbf{Split} & \textbf{Model} & \textbf{\begin{tabular}[c]{@{}l@{}}Train\\ Acc.\end{tabular}} & \textbf{\begin{tabular}[c]{@{}l@{}}Test \\ Acc.\end{tabular}} & \textbf{F1-Score} & \textbf{Prec.} & \textbf{Recall} & \textbf{\begin{tabular}[c]{@{}l@{}}F1-Score\\ Rain\end{tabular}} & \textbf{\begin{tabular}[c]{@{}l@{}}F1-Score\\ Snow\end{tabular}} & \textbf{\begin{tabular}[c]{@{}l@{}}F1-Score\\ Dust\end{tabular}} & \textbf{\begin{tabular}[c]{@{}l@{}}F1-Score\\Fog\end{tabular}} \\ \hline
\multirow{10}*{80} & Xception & 99.69 & 71.88 & 0.73 & 0.72 & 0.72 & 0.62 & \textbf{0.93} & 0.77 & 0.56 \\
 & VGG-16 & 99.69 & \textbf{78.75} & \textbf{0.79} & \textbf{0.79} & \textbf{0.79} & 0.68 & 0.9 & \textbf{0.88} & \textbf{0.71} \\
 & VGG-19 & 99.53 & 70.63 & 0.7 & 0.71 & 0.7 & 0.55 & 0.82 & 0.79 & 0.63 \\
 & ResNet50 & 41.56 & 45.63 & 0.5 & 0.49 & 0.45 & 0.45 & 0.73 & 0.34 & 0.28 \\
 & MobileNet & 99.69 & 78.12 & \textbf{0.79} & 0.78 & 0.78 & \textbf{0.73} & 0.91 & 0.87 & 0.63 \\
 & DenseNet & 99.22 & 77.50 & 0.78 & 0.76 & 0.77 & 0.73 & 0.88 & 0.83 & 0.63 \\
 & InceptionV3 & 99.69 & 70.63 & 0.72 & 0.71 & 0.71 & 0.68 & 0.89 & 0.72 & 0.55 \\
 & MobileNetV2 & 99.69 & 73.12 & 0.73 & 0.73 & 0.72 & 0.63 & 0.86 & 0.79 & 0.62 \\
 & EfficientNetV2S & 75.63 & 48.12 & 0.49 & 0.49 & 0.46 & 0.32 & 0.72 & 0.26 & 0.55 \\
 & ConvNeXtSmall & 61.56 & 51.25 & 0.41 & 0.41 & 0.4 & 0.31 & 0.59 & 0.3 & 0.42 \\ \hline
\multirow{10}*{50} & Xception & 99.75 & 72.50 & 0.73 & 0.72 & 0.72 & 0.68 & 0.87 & 0.73 & 0.61 \\
 & VGG-16 & 99.75 & 69.38 & 0.7 & 0.69 & 0.69 & 0.61 & 0.88 & 0.78 & 0.5 \\
 & VGG-19 & 99.75 & 70.00 & 0.7 & 0.7 & 0.7 & 0.59 & 0.84 & 0.78 & 0.58 \\
 & ResNet50 & 58.75 & 55.62 & 0.51 & 0.51 & 0.49 & 0.53 & 0.7 & 0.25 & 0.46 \\
 & MobileNet & 99.75 & 73.12 & 0.74 & 0.73 & 0.73 & 0.65 & 0.87 & 0.8 & 0.61 \\
 & DenseNet & 99.75 & 74.37 & 0.75 & 0.74 & 0.74 & 0.67 & 0.89 & 0.81 & 0.6 \\
 & InceptionV3 & 99.75 & 63.75 & 0.62 & 0.63 & 0.62 & 0.63 & 0.77 & 0.63 & 0.42 \\
 & MobileNetV2 & 99.75 & 70.63 & 0.7 & 0.71 & 0.7 & 0.65 & 0.85 & 0.7 & 0.59 \\
 & EfficientNetV2S & 74.50 & 47.50 & 0.46 & 0.47 & 0.45 & 0.49 & 0.73 & 0.18 & 0.41 \\
 & ConvNeXtSmall & 65.25 & 50.63 & 0.46 & 0.47 & 0.45 & 0.49 & 0.73 & 0.18 & 0.41 \\ \hline
\multirow{10}*{20} & Xception & \textbf{100.00} & 67.50 & 0.67 & 0.68 & 0.67 & 0.62 & 0.82 & 0.7 & 0.54 \\
 & VGG-16 & \textbf{100.00} & 65.00 & 0.66 & 0.66 & 0.66 & 0.6 & 0.76 & 0.68 & 0.6 \\
 & VGG-19 & \textbf{100.00} & 62.50 & 0.62 & 0.62 & 0.62 & 0.54 & 0.73 & 0.64 & 0.58 \\
 & ResNet50 & 61.87 & 48.75 & 0.5 & 0.47 & 0.47 & 0.52 & 0.57 & 0.26 & 0.52 \\
 & MobileNet & \textbf{100.00} & 73.12 & 0.74 & 0.73 & 0.73 & 0.67 & 0.86 & 0.78 & 0.6 \\
 & DenseNet & \textbf{100.00} & 71.88 & 0.71 & 0.71 & 0.71 & 0.65 & 0.89 & 0.76 & 0.54 \\
 & InceptionV3 & \textbf{100.00} & 56.88 & 0.62 & 0.57 & 0.53 & 0.56 & 0.76 & 0.62 & 0.17 \\
 & MobileNetV2 & \textbf{100.00} & 63.13 & 0.66 & 0.63 & 0.62 & 0.55 & 0.84 & 0.62 & 0.45 \\
 & EfficientNetV2S & 85.62 & 48.12 & 0.45 & 0.45 & 0.42 & 0.52 & 0.58 & 0.29 & 0.31 \\
 & ConvNeXtSmall & 68.75 & 44.37 & 0.59 & 0.42 & 0.37 & 0.49 & 0.52 & 0.05 & 0.44\\ \hline
\end{tabular} 
\caption{\textbf{Weather Classification Benchmark} (Learning with Limited Data)  with Benchmark Models (Xception \cite{chollet2017xception},	VGG16 \cite{simonyan2014very},	VGG19 \cite{simonyan2014very},	ResNet50 \cite{he2016deep},	MobileNet\cite{howard2017mobilenets},	DenseNet\cite{huang2017densely},	InceptionV3 \cite{szegedy2016rethinking},	MobileNetV2\cite{howard2017mobilenets},	EfficientNetV2S \cite{tan2021efficientnetv2}, ConvNeXtSmall\cite{liu2022convnet}). As expected, the models trained with 80-20 split (greatest training set) deliver the best results.}
\label{tab1}
\end{table*}

\section{Methodology}

\subsection{Datasets}
We employ the DAWN Dataset \cite{kenk2020dawn} and the WEDGE dataset \cite{aa} to test the efficacy of our strategy. The DAWN dataset is a 1000 image object detection dataset that includes traffic imagery in bad weather like rain,fog, dust and snow. WEDGE is a synthetic dataset that employs the DALL-E  2 model  \cite{ramesh2021zero,ramesh2022hierarchical} with prompts encompassing 16 weather and season conditions with a focus on autonomous vehicle scenarios.
 It features images captured during severe weather, such as rain, snow, fog, and sandstorms (Refer Figure \ref{fig1}).

\subsection{Proposed t-RAIN Algorithm}

In limited data settings, generalization capabilities are naturally limited by the number of examples seen by the classifier. However, with the development of large-scale vision-language models (stable diffusion, DALL-E), access to unlimited synthetic datasets has become much easier. We propose an algorithm that can leverage classical classifiers with synthetically generated data to provide generalization capabilities even in limited data settings ($<160$ images).
\begin{algorithm}
\begin{algorithmic}[2]
\REQUIRE{Randomly synthesized unlabelled dataset $Q_t$}\newline
{Labelled real training dataset $Q$ with C classes}
\FOR{$i\in Train-Set \;  Size (or \; B \; iterations)$}
  \STATE Sample data point $X_i$ at random
     \STATE $\phi_i \; \gets \;  Oracle( X_i, C_i) $
     \STATE $ Q_{C_i} \; \gets \;  \phi_i$
     \label{line:mostimp1}
   \ENDFOR
\RETURN $Q$
\newline
\newline
\textit{Sub-Program: Oracle for Mapping Sim2Real Samples}
\newline
\REQUIRE{Sample $X_i$, Class $C_i$}\newline
{Labelled synthetic dataset $S$ with C classes (extracted from prompts)}
\ENSURE{$C_i \in C$}
\STATE $j \gets 1$ 
\FOR{ $j \leq  \eta  $}
  \STATE Sample synthetic data point $Q_{j}$ at random
  \STATE$\; Sc_{j}  \gets Cosine\_Similarity(Class(Q_{j}),C_i)$
   \STATE$\psi_{j} \gets Q_{j}$
   
     \label{line:mostimp}
\ENDFOR
 \STATE SORT $ \; \psi \; by \; Sc_{j} $

  \STATE FILTER $ \;   \phi_i <- \psi[\eta - \beta: \eta] $ \newline
 $(\beta=210 \; maximum \; values \; here ) $  
\RETURN $\phi_i$
\end{algorithmic}
\caption{t-RAIN Algorithm}
\label{algo:first}
\end{algorithm}

\begin{equation}
    cos( x,  y) = \frac { x \cdot y}{|| x|| \cdot || y||}
    \label{eq1}
\end{equation}

The algorithm works by mapping similarity between the source distribution classes (C) and target synthetic classes (C) to sample relevant images and up-sample for each class. The technique currently performs uniform weighting but can be extended to weighted sampling in the future works. Once the source class and target class are compared (this can be between class and prompt keywords as well), we filter the images with greatest similarity (currently cosine similarity \ref{eq1}).

\begin{figure}[h]
    \centering
    \includegraphics[width=0.48\textwidth,height=5cm]{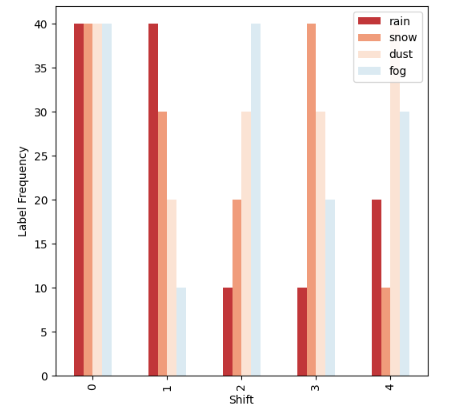}
    \caption{\textbf{Label Shift Simulation}: Shifts 0,1,2,3,4 correspond to the simulated No-Shift, Rain, Fog, Snow, Dust Shifts' target label distributions. }
    \label{fig3}
\end{figure}

\begin{figure*}[h!]
    \centering
    \includegraphics[width=0.98\textwidth,height=8cm]{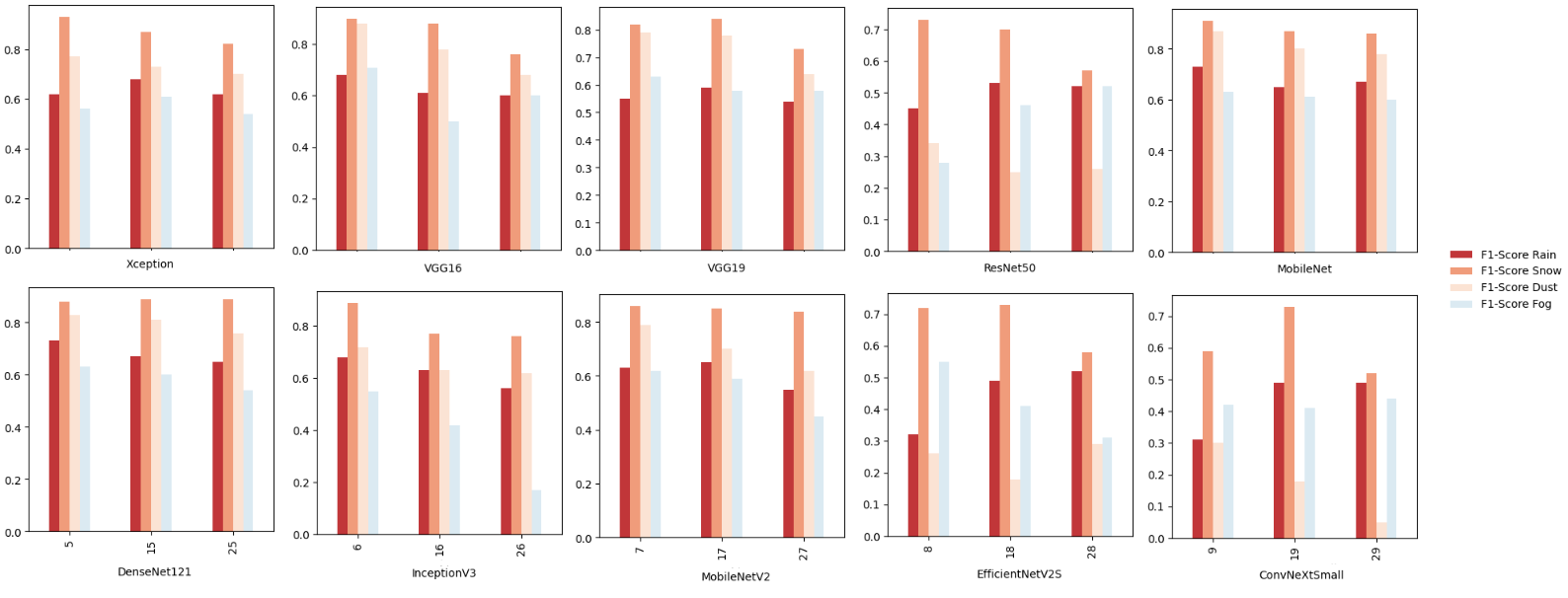}
    \caption{\textbf{How well do models differenciate between weather? } Evaluation of class-level performance of models under 80-20, 50-50 and 20-80 train-test splits using F1-Score Metric on DAWN dataset \cite{kenk2020dawn}.}
    \label{fig4}
\end{figure*}

\begin{table*}[]
\begin{centering}
    \begin{tabular}{llllllllllll} \hline
\textbf{Split} & \textbf{Shift} & \textbf{M1} & \textbf{M2} & \textbf{M3} & \textbf{M4} & \textbf{M5} & \textbf{M6} & \textbf{M7} & \textbf{M8} & \textbf{M9} & \textbf{M10} \\ \hline

\multirow{5}{*}{80} & 1 & 71 & 78 & 70 & 45 & 78 & 77 & 70 & 73 & 48 & 51 \\
 & 2 & 70 & 77 & 68 & 56 & 81 & 79 & \textbf{73} & 72 & 42 & 49 \\
 & 3 & 71 & 81 & 71 & 38 & 79 & 75 & 66 & 73 & \textbf{56} & 54 \\
 & 4 & 75 & \textbf{83} & \textbf{76} & 53 & \textbf{81} & \textbf{81} & 73 & 78 & 53 & \textbf{57} \\ 
 & 5 & 67 & 80 & 69 & 35 & 78 & 74 & 69 & 75 & 41 & 44 \\ \hline
\multirow{5}{*}{50} & 1 & 72 & 69 & 70 & 55 & 73 & 74 & 63 & 70 & 47 & 50 \\
 & 2 & \textbf{77} & 75 & 74 & 59 & 76 & 76 & 70 & 76 & 46 & 50 \\
 & 3 & 73 & 68 & 70 & 52 & 71 & 74 & 59 & 69 & 43 & 52 \\
 & 4 & 75 & 77 & 76 & 59 & 77 & 81 & 71 & \textbf{79} & 48 & 53 \\ 
 & 5 & 67 & 68 & 66 & 46 & 69 & 71 & 59 & 69 & 37 & 43 \\\hline
\multirow{5}{*}{20} & 1 & 67 & 65 & 62 & 48 & 73 & 71 & 56 & 63 & 48 & 44 \\
 & 2 & 70 & 68 & 65 & 51 & 77 & 74 & 68 & 66 & 53 & 47 \\
 & 3 & 67 & 68 & 64 & 48 & 68 & 70 & 44 & 61 & 43 & 35 \\
 & 4 & 74 & 69 & 67 & 45 & 77 & 77 & 58 & 73 & 50 & 35 \\
 & 5 & 61 & 66 & 64 & 43 & 67 & 69 & 50 & 63 & 35 & 36 \\ \hline
\multirow{5}{*}{t-RAIN} & 1 & 70 & 68 & 65 & 55 & 71 & 74 & 64 & 66 & 43 & 42 \\
 & 2 & 70 & 71 & 62 & \textbf{61} & 74 & 76 & 66 & 71 & 42 & 38 \\
 & 3 & 70 & 69 & 65 & 50 & 72 & 72 & 60 & 64 & 47 & 43 \\
 & 4 & 72 & 73 & 69 & 56 & 80 & 78 & 68 & 67 & 43 & 38 \\
 & 5 & 66 & 67 & 62 & 50 & 68 & 70 & 58 & 63 & 38 & 39 \\ \hline
\end{tabular}

\caption{\textbf{Weather Classification Benchmark: }Test Accuracy of Benchmark Models M1 to M10 from Left to Right (Xception \cite{chollet2017xception},	VGG16 \cite{simonyan2014very},	VGG19 \cite{simonyan2014very},	ResNet50 \cite{he2016deep},	MobileNet\cite{howard2017mobilenets},	DenseNet\cite{huang2017densely},	InceptionV3 \cite{szegedy2016rethinking},	MobileNetV2\cite{howard2017mobilenets},	EfficientNetV2S \cite{tan2021efficientnetv2}, ConvNeXtSmall\cite{liu2022convnet}) After Label Shift 
Shift 1 : None, Shift 2: Rain , Shift 3: Fog , Shift 4: Snow , Shift 5: Dust  on DAWN dataset \cite{kenk2020dawn}. }
\label{tab2}
\end{centering}

\end{table*}

\begin{figure}[h]
    \centering
    \includegraphics[width=0.48\textwidth,height=7cm]{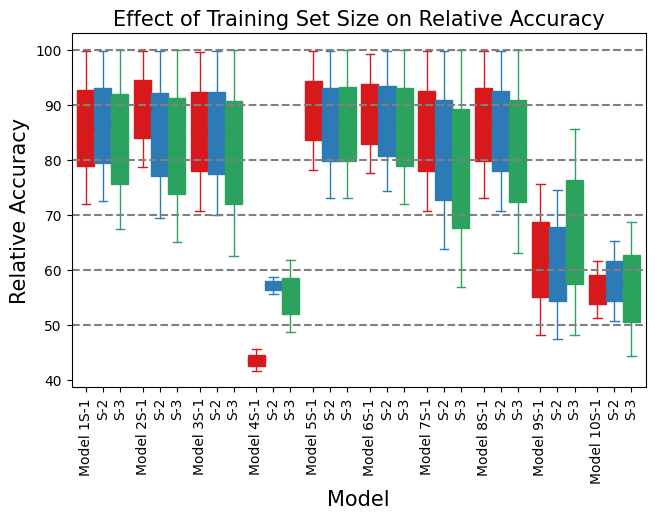}
    \caption{\textbf{Size matters!}: Effect of training set size on model performance, S-1, S-2, S-3 represents the 80-20 (red), 50-50 (blue) and 20-80 (green) train-test split variations of the trained models  on DAWN dataset \cite{kenk2020dawn}.}
    \label{figx}
\end{figure}

\begin{table}[]
\begin{center}
    \begin{tabular}{lll}
\hline
\textbf{Weather/Data} & \textbf{DAWN} & \textbf{WEDGE} \\ \hline
Rain & 73.5 & 29.7 \\
Snow & \textbf{82.69} & 22.34 \\
Dust & 59.66 & 60.89 \\
Fog & 73.32 & \textbf{62.31} \\ \hline
\end{tabular}

\end{center}
\caption{\textbf{Pedestrian detection in adverse weather}: We observe that best detection performance is under Real Snow Conditions with \textbf{82.69 AP} and Synthetic Fog Conditions  with \textbf{62.31 AP} when evaluated on DAWN and WEDGE datasets. The number of images used for evaluation was 766 and 810 respectively. The model for detection is FasterRCNN with Resnet 50 Backbone pre-trained on COCO images \cite{lin2014microsoft}.}
\label{tab4}
\end{table}

We return this set through the oracle until all classes are mapped and sufficient samples  (determined by hyper-parameter $\beta$) from the target dataset (of size $\eta$) are satisfactorily generated. The filtered target samples from class with closest class similarity to source sample ($X_i$) are the augmentation set.

\section{Experiments}

The experimentation procedure was carried out in the following steps:
\begin{enumerate}
    \item Training of set of classifiers on DAWN dataset targeted for optimal classification accuracy.
    \item Performance evaluation on smaller training sets (80-50-20 splits) and robustness evaluation.
    \item Simulation of 4 label shift scenarios and comparison with uniform label distribution.
    \item Pedestrian detection under 4 adverse conditions for 2 datasets: Real (DAWN) and Synthetic (WEDGE). 
    
\end{enumerate}

\begin{figure*}[h!]
    \centering
    \includegraphics[width=0.98\textwidth,height=5cm]{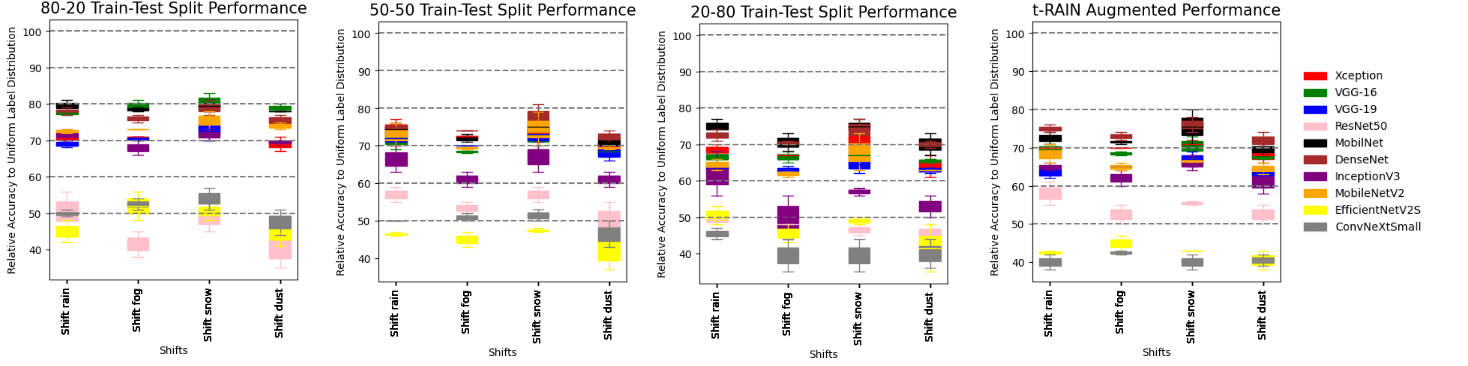}
    \caption{\textbf{Understanding multi-weather robustness: }Relative Performance of classification models under different label shift and training data distributions  on DAWN dataset \cite{kenk2020dawn}.}
    \label{figxy}
\end{figure*}

\section{Results and Discussion}

\subsection{Model Generalization in Limited Data Settings}
As visible in Table \ref{tab1} and Figure \ref{figx}, the model performance drops significantly when restricted to limited data environments as expected. The models begin overfitting on training set and are unable to generalize to multi-weather conditions. EfficientNetV2S and ConvNeXtSmall  (the weakest learning models) have shown improvement on limited data settings, indicating the pseudo-generalization capabilities of weak learners, albeit extremely poor performance due to underfitting. 
\subsection{Label Shift Generalization}

We simulate label shift by boosting one class at a time (Figure \ref{fig3}) and a simple random affine transformation on the remaining set sizes. The goal is to observe shift when specific weather conditions dominate the target distribution.

\subsection{Benchmark Comparison}

\begin{figure*}[h!]
    \centering
    \includegraphics[width=\textwidth,height=4cm]{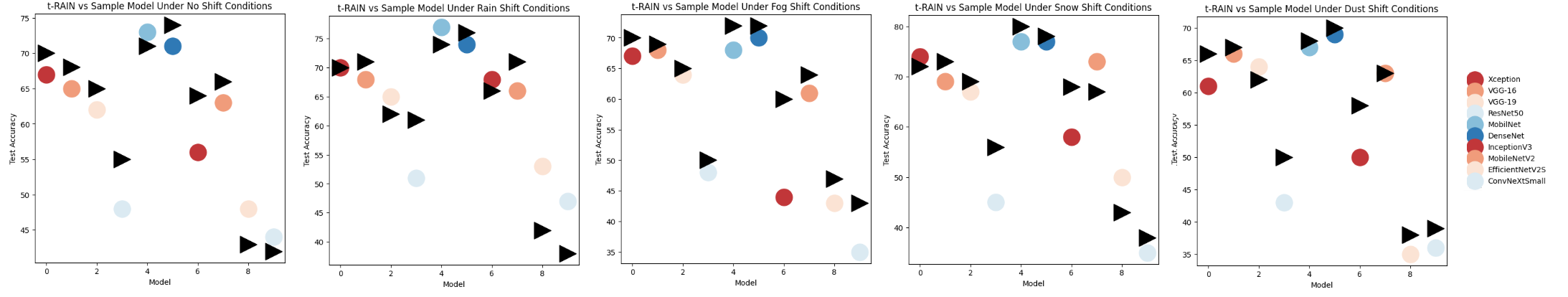}
    \caption{\textbf{Contribution of t-RAIN to Model Generalization}: We demonstrate model performance under 5 shift conditions: No shift, Rain, Fog, Snow, Dust in the above 5 figures from Left to Right. The black arrows indicate the test accuracy of t-RAIN algorithm and in-line coloured circles represent individual models. Whenever the black arrow appears above the circle, the t-RAIN outperforms limited data benchmark  on DAWN dataset \cite{kenk2020dawn}.}
    \label{fig6}
\end{figure*}

\begin{figure*}[h!]
    \centering
    \includegraphics[width=0.5\textwidth,height=9cm]{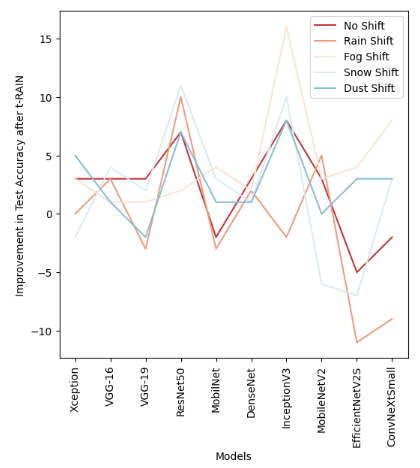}
    \caption{\textbf{Performance Evaluation}: The lines demonstrate model-wise performance differences as measured by relative accuracy between limited data benchmarks and proposed t-RAIN algorithm under all 5 shift conditions. Fog shift appears to have the most dramatic improvement in test-time performance on DAWN dataset \cite{kenk2020dawn}.}
    \label{fig7}
\end{figure*}

\begin{figure}[h!]
    \centering
    \includegraphics[width=0.48\textwidth,height=7cm]{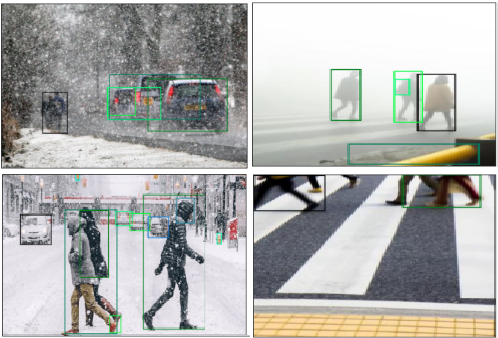}
    \caption{\textbf{Finding people in all seasons:} Pedestrian Detection in Real (DAWN) and Synthetic (WEDGE) imagery (left to right).}
    \label{fig8}
\end{figure}

We can see from Table \ref{tab2} and Figures \ref{fig4},\ref{fig6},\ref{fig7}, some model-specific trends:
\begin{enumerate}
    \item  Xception model which had fit very well on snow class, performed better when the snow class was positively biased in the target distribution. Similarly it performs worse when exposed to other shifts which it had learned poorly. Upon applying t-RAIN it shows 3 \% increase in accuracy for no-shift and fog shift conditions. It shows 5\% increase in accuracy for dust-shift attacks.
     \item VGG-16 is one of the stronger learning models with higher performance values. It performs similar on all classes and attains high performance under all shifts, especially snow. When synthetic data is sampled via t-RAIN, it consistently outperforms baseline VGG-16 with 1-4\% increases. This may point towards better generalization of stronger models due to efficient learning of complex representations from the extended input space. 
      \item VGG-19 shows similar trends to VGG-16 but with 3\% under-performing margin for rain and dust shifts.
       \item  ResNet50 is one of the weakest learners with second-last performance in most cases. It generalizes better in limited data due to under-fitting and shows mediocre results which improve marginally for snow shifts. t-RAIN dramatically improves the ResNet performance on all shifts by 2-11 \% accuracy. This is an important result, as we observe that variability in data can boost both strong and weak learners, with greater effects on weak learners.
        \item EfficientNetV2S is a significantly weak learner with worst performance on dust shift. t-RAIN is only able to boost performance by 3-4 \% on dust and fog shifts.
             \item ConvNextSmall is also one of the more poorly performing models with almost zero robustness to weather conditions like dust shift. Although t-RAIN improves robustness under all conditions by 3 - 8 \% except rain and no-shift the model suffers under limited data constraints and plummets to bottom rank.
        \item MobileNet features good performance and fast training. It is not robust to fog corruptions but otherwise provides reasonable results with 1-4\% boosts  in majority classes and worse under rain and no-shifts.
         \item DenseNet features similar trends with baseline performance mainly for snow shifts. It features the most common boost of 1-3 \% over all shifts uniformly.
            \item MobileNetV2 features good performance and fast training. The model improved with t-RAIN under rain and fog shifts by 5 \% and 3 \% respectively.        
          \item Inception V3 is an average learner but suffers adversely from fog shift. Adding t-RAIN to such models significantly improves the performance which is a remarkable result. There is 8 and 16\% increase in test accuracy after using t-RAIN to improve no-shift and  fog robustness in Inception V3.Adding t-RAIN improves the performance the most dramatically out of all the other results with 16\% increase in test accuracy here under fog shift which is a remarkable result. 
        
\end{enumerate}

Some general observations include snow being one of the easiest weather classes to recognize due to significant distinguishing characteristics. Models suffer from easier evaluation when snow is considered as one of the evaluation classes. For true robustness evaluation such classes should be held out and only measured as sanity checks and not robustness measures. 
As visible in Figure \ref{fig6}, the t-RAIN algorithm is able to improve generalization , for all learners from strong learners like VGG-16  to weak learners like  EfficientNetV2S and outperforms the performance on limited data with synthetic augmentation. The averaged improvement across all 5 shifts are 2.1, -0.8,4.4,1.9, 2.7 \% respectively. One interesting result presents a question of why t-RAIN boosts specific class -shifts performance inspite of underlying uniform distribution and uniform augmentation. This could potentially be attributed to special variations and robustness introduced by the synthetic data which helped gain generalization capabilities beyond the source distribution.

\subsection{Pedestrian Detection}

We use the earlier analysis as a prior to analyze performance of models in detection pedestrians under anomalous weather conditions. As visible in Table \ref{tab4} and Figure \ref{fig8}, the best detection performance is under Real Snow Conditions with \textbf{82.69 AP} and Synthetic Fog Conditions  with \textbf{62.31 AP} when evaluated on DAWN and WEDGE datasets. Intuitively another direction we explore the adversarial difficulty of each weather condition for detecting pedestrians. In real data, dust weather appears to obstruct vision for pedestrian detectors the most whereas in synthetic data, snow appears to obstruct vision the most. This is a surprising result as snow is the easiest weather in real data, which is the opposite in synthetic data. Due to this adversary, we can understand now why t-RAIN improves generalization capability of 7/10 models under snow shift with a maximum increase of 10\% accuracy. However, conversely, ease in detection like fog conditions does not imply ineffectiveness of t-RAIN. t-RAIN improves generalization capability of 10/10 models under fog shift with a maximum increase of 16\% accuracy. Further analysis can be done in the direction of discovering distribution shift induced by synthetic generation. This analysis was not performed in the scope of this study, but there can be multiple possible underlying factors for the results reported in Table \ref{tab4}. Firstly, we can see that generated humans in  synthetic data appear out-of-distribution due to the trust and security layers implemented for privacy concerns and obscuration. Thus evaluation of pedestrian detection across Sim2Real data is not actually an informative indicator of generative accuracy but may hint towards undiscovered generative anomalies. Models that work well on real-data and poorly on synthetic data could be suffering due to (a) Real2Sim gap (b) Beneficial adversarial robustness or (c) Harmful generative anomalies that do not help real-world models. Identifying the exact cause for performance shift is an interesting challenge that we propose for future works.

\section{Conclusion}

Better overall test performance may not always signify better multi-weather generalization, but could be attributed to underlying factors like unseen target distribution shifts. Models may have possibly rote-learned specific classes and still go unseen as bad learners due to convenient boosts in model performance due to label shifts. Weak learners also show  pseudo-generalization capabilities which are usually too small in magnitude  and misleading to be considered significant. Leveraging weaker learners through ensemble methods can be explored in the future scope of this study.  Through this small-scale study, we were able to uncover many insights on the fundamental problems with multi-weather robustness as an extension on the label shift and generalization problems of benchmark classification models.

The applications of this study mainly apply to autonomous perception in unsupervised settings, where model robustness is difficult to evaluate and target distributions are often skewed. They can extend to all real world scenarios like medical image analysis, species classification etc that showcase out-of-distribution examples and variable label distributions. Given unlabelled target data, one can attain reasonable results if model is robust to label shift on uniform source label distribution. One might also attempt to predict unlabelled target weather distribution upto a certain confidence using a well-trained model from this work. Another application can include using weather-classification labels as a prior for downstream computer vision tasks like specialized image denoising specific to the weather condition. We consider the integration of large-scale generative models into our study as an example of improvement on classical data collection methods with novel architectures for better generalization. In the future work, we would like to put forward better methods for overcoming label shift vulnerability and weather-specific methods for robust all-weather vision extended to unsupervised settings.

{\small
\bibliographystyle{ieee_fullname}
\bibliography{egbib}

\begin{thebibliography}{10}\itemsep=-1pt

\bibitem{buades2005non}
Antoni Buades, Bartomeu Coll, and J-M Morel.
\newblock A non-local algorithm for image denoising.
\newblock In {\em 2005 IEEE computer society conference on computer vision and
  pattern recognition (CVPR'05)}, volume~2, pages 60--65. Ieee, 2005.

\bibitem{buades2005review}
Antoni Buades, Bartomeu Coll, and Jean-Michel Morel.
\newblock A review of image denoising algorithms, with a new one.
\newblock {\em Multiscale modeling \& simulation}, 4(2):490--530, 2005.

\bibitem{caesar2020nuscenes}
Holger Caesar, Varun Bankiti, Alex~H Lang, Sourabh Vora, Venice~Erin Liong,
  Qiang Xu, Anush Krishnan, Yu Pan, Giancarlo Baldan, and Oscar Beijbom.
\newblock nuscenes: A multimodal dataset for autonomous driving.
\newblock In {\em Proceedings of the IEEE/CVF conference on computer vision and
  pattern recognition}, pages 11621--11631, 2020.

\bibitem{cai2016unified}
Zhaowei Cai, Quanfu Fan, Rogerio~S Feris, and Nuno Vasconcelos.
\newblock A unified multi-scale deep convolutional neural network for fast
  object detection.
\newblock In {\em Computer Vision--ECCV 2016: 14th European Conference,
  Amsterdam, The Netherlands, October 11--14, 2016, Proceedings, Part IV 14},
  pages 354--370. Springer, 2016.

\bibitem{chollet2017xception}
Fran{\c{c}}ois Chollet.
\newblock Xception: Deep learning with depthwise separable convolutions.
\newblock In {\em Proceedings of the IEEE conference on computer vision and
  pattern recognition}, pages 1251--1258, 2017.

\bibitem{dalal2005histograms}
Navneet Dalal and Bill Triggs.
\newblock Histograms of oriented gradients for human detection.
\newblock In {\em 2005 IEEE computer society conference on computer vision and
  pattern recognition (CVPR'05)}, volume~1, pages 886--893. Ieee, 2005.

\bibitem{dendorfer2019cvpr19}
Patrick Dendorfer, Hamid Rezatofighi, Anton Milan, Javen Shi, Daniel Cremers,
  Ian Reid, Stefan Roth, Konrad Schindler, and Laura Leal-Taixe.
\newblock Cvpr19 tracking and detection challenge: How crowded can it get?
\newblock {\em arXiv preprint arXiv:1906.04567}, 2019.

\bibitem{dendorfer2020mot20}
Patrick Dendorfer, Hamid Rezatofighi, Anton Milan, Javen Shi, Daniel Cremers,
  Ian Reid, Stefan Roth, Konrad Schindler, and Laura Leal-Taix{\'e}.
\newblock Mot20: A benchmark for multi object tracking in crowded scenes.
\newblock {\em arXiv preprint arXiv:2003.09003}, 2020.

\bibitem{ess2008mobile}
Andreas Ess, Bastian Leibe, Konrad Schindler, and Luc Van~Gool.
\newblock A mobile vision system for robust multi-person tracking.
\newblock In {\em 2008 IEEE Conference on Computer Vision and Pattern
  Recognition}, pages 1--8. IEEE, 2008.

\bibitem{w3}
Saurabh Garg, Sivaraman Balakrishnan, Zico Kolter, and Zachary Lipton.
\newblock Ratt: Leveraging unlabeled data to guarantee generalization.
\newblock In Marina Meila and Tong Zhang, editors, {\em Proceedings of the 38th
  International Conference on Machine Learning}, volume 139 of {\em Proceedings
  of Machine Learning Research}, pages 3598--3609. PMLR, 18--24 Jul 2021.

\bibitem{w1}
Saurabh Garg, Sivaraman Balakrishnan, Zachary~C. Lipton, Behnam Neyshabur, and
  Hanie Sedghi.
\newblock Leveraging unlabeled data to predict out-of-distribution performance,
  2022.

\bibitem{w2}
Saurabh Garg, Yifan Wu, Sivaraman Balakrishnan, and Zachary~C. Lipton.
\newblock A unified view of label shift estimation.
\newblock In {\em Proceedings of the 34th International Conference on Neural
  Information Processing Systems}, NIPS'20, Red Hook, NY, USA, 2020. Curran
  Associates Inc.

\bibitem{geiger2012we}
Andreas Geiger, Philip Lenz, and Raquel Urtasun.
\newblock Are we ready for autonomous driving? the kitti vision benchmark
  suite.
\newblock In {\em 2012 IEEE conference on computer vision and pattern
  recognition}, pages 3354--3361. IEEE, 2012.

\bibitem{he2016deep}
Kaiming He, Xiangyu Zhang, Shaoqing Ren, and Jian Sun.
\newblock Deep residual learning for image recognition.
\newblock In {\em Proceedings of the IEEE conference on computer vision and
  pattern recognition}, pages 770--778, 2016.

\bibitem{howard2017mobilenets}
Andrew~G Howard, Menglong Zhu, Bo Chen, Dmitry Kalenichenko, Weijun Wang,
  Tobias Weyand, Marco Andreetto, and Hartwig Adam.
\newblock Mobilenets: Efficient convolutional neural networks for mobile vision
  applications.
\newblock {\em arXiv preprint arXiv:1704.04861}, 2017.

\bibitem{huang2017densely}
Gao Huang, Zhuang Liu, Laurens Van Der~Maaten, and Kilian~Q Weinberger.
\newblock Densely connected convolutional networks.
\newblock In {\em Proceedings of the IEEE conference on computer vision and
  pattern recognition}, pages 4700--4708, 2017.

\bibitem{kenk2020dawn}
Mourad~A Kenk and Mahmoud Hassaballah.
\newblock Dawn: vehicle detection in adverse weather nature dataset.
\newblock {\em arXiv preprint arXiv:2008.05402}, 2020.

\bibitem{khurana2021detecting}
Tarasha Khurana, Achal Dave, and Deva Ramanan.
\newblock Detecting invisible people.
\newblock In {\em Proceedings of the IEEE/CVF international conference on
  computer vision}, pages 3174--3184, 2021.

\bibitem{lin2014microsoft}
Tsung-Yi Lin, Michael Maire, Serge Belongie, James Hays, Pietro Perona, Deva
  Ramanan, Piotr Doll{\'a}r, and C~Lawrence Zitnick.
\newblock Microsoft coco: Common objects in context.
\newblock In {\em Computer Vision--ECCV 2014: 13th European Conference, Zurich,
  Switzerland, September 6-12, 2014, Proceedings, Part V 13}, pages 740--755.
  Springer, 2014.

\bibitem{liu2022convnet}
Zhuang Liu, Hanzi Mao, Chao-Yuan Wu, Christoph Feichtenhofer, Trevor Darrell,
  and Saining Xie.
\newblock A convnet for the 2020s.
\newblock In {\em Proceedings of the IEEE/CVF Conference on Computer Vision and
  Pattern Recognition}, pages 11976--11986, 2022.

\bibitem{mao2017can}
Jiayuan Mao, Tete Xiao, Yuning Jiang, and Zhimin Cao.
\newblock What can help pedestrian detection?
\newblock In {\em Proceedings of the IEEE Conference on Computer Vision and
  Pattern Recognition}, pages 3127--3136, 2017.

\bibitem{marathe2022restorex}
Aboli Marathe, Pushkar Jain, Rahee Walambe, and Ketan Kotecha.
\newblock Restorex-ai: A contrastive approach towards guiding image restoration
  via explainable ai systems.
\newblock In {\em Proceedings of the IEEE/CVF Conference on Computer Vision and
  Pattern Recognition}, pages 3030--3039, 2022.

\bibitem{aa}
Aboli Marathe, Deva Ramanan, Rahee Walambe, and Ketan Kotecha.
\newblock {WEDGE}: A multi-weather autonomous driving dataset built from
  generative vision-language models.
\newblock {\em arXiv preprint arXiv:2305.07528}, 2023.

\bibitem{marathe2021evaluating}
Aboli Marathe, Rahee Walambe, and Ketan Kotecha.
\newblock Evaluating the performance of ensemble methods and voting strategies
  for dense 2d pedestrian detection in the wild.
\newblock In {\em Proceedings of the IEEE/CVF International Conference on
  Computer Vision}, pages 3575--3584, 2021.

\bibitem{marathe2022rain}
Aboli Marathe, Rahee Walambe, Ketan Kotecha, and Deepak~Kumar Jain.
\newblock In rain or shine: Understanding and overcoming dataset bias for
  improving robustness against weather corruptions for autonomous vehicles.
\newblock {\em arXiv preprint arXiv:2204.01062}, 2022.

\bibitem{motwani2004survey}
Mukesh~C Motwani, Mukesh~C Gadiya, Rakhi~C Motwani, and Frederick~C Harris.
\newblock Survey of image denoising techniques.
\newblock In {\em Proceedings of GSPX}, volume~27, pages 27--30. Proceedings of
  GSPX, 2004.

\bibitem{ms}
Valentina Mușat, Ivan Fursa, Paul Newman, Fabio Cuzzolin, and Andrew Bradley.
\newblock Multi-weather city: Adverse weather stacking for autonomous driving.
\newblock In {\em Proceedings of the IEEE/CVF International Conference on
  Computer Vision}, pages 2906--2915, 2021.

\bibitem{ramesh2022hierarchical}
Aditya Ramesh, Prafulla Dhariwal, Alex Nichol, Casey Chu, and Mark Chen.
\newblock Hierarchical text-conditional image generation with clip latents.
\newblock {\em arXiv preprint arXiv:2204.06125}, 2022.

\bibitem{ramesh2021zero}
Aditya Ramesh, Mikhail Pavlov, Gabriel Goh, Scott Gray, Chelsea Voss, Alec
  Radford, Mark Chen, and Ilya Sutskever.
\newblock Zero-shot text-to-image generation.
\newblock In {\em International Conference on Machine Learning}, pages
  8821--8831. PMLR, 2021.

\bibitem{ren2015faster}
Shaoqing Ren, Kaiming He, Ross Girshick, and Jian Sun.
\newblock Faster r-cnn: Towards real-time object detection with region proposal
  networks.
\newblock {\em Advances in neural information processing systems}, 28, 2015.

\bibitem{saikia2021improving}
Tonmoy Saikia, Cordelia Schmid, and Thomas Brox.
\newblock Improving robustness against common corruptions with frequency biased
  models.
\newblock In {\em Proceedings of the IEEE/CVF International Conference on
  Computer Vision}, pages 10211--10220, 2021.

\bibitem{sermanet2013pedestrian}
Pierre Sermanet, Koray Kavukcuoglu, Soumith Chintala, and Yann LeCun.
\newblock Pedestrian detection with unsupervised multi-stage feature learning.
\newblock In {\em Proceedings of the IEEE conference on computer vision and
  pattern recognition}, pages 3626--3633, 2013.

\bibitem{sigal2004tracking}
Leonid Sigal, Sidharth Bhatia, Stefan Roth, Michael~J Black, and Michael Isard.
\newblock Tracking loose-limbed people.
\newblock In {\em Proceedings of the 2004 IEEE Computer Society Conference on
  Computer Vision and Pattern Recognition, 2004. CVPR 2004.}, volume~1, pages
  I--I. IEEE, 2004.

\bibitem{simonyan2014very}
Karen Simonyan and Andrew Zisserman.
\newblock Very deep convolutional networks for large-scale image recognition.
\newblock {\em arXiv preprint arXiv:1409.1556}, 2014.

\bibitem{szegedy2016rethinking}
Christian Szegedy, Vincent Vanhoucke, Sergey Ioffe, Jon Shlens, and Zbigniew
  Wojna.
\newblock Rethinking the inception architecture for computer vision.
\newblock In {\em Proceedings of the IEEE conference on computer vision and
  pattern recognition}, pages 2818--2826, 2016.

\bibitem{tan2021efficientnetv2}
Mingxing Tan and Quoc Le.
\newblock Efficientnetv2: Smaller models and faster training.
\newblock In {\em International conference on machine learning}, pages
  10096--10106. PMLR, 2021.

\bibitem{viola2005detecting}
Paul Viola, Michael~J Jones, and Daniel Snow.
\newblock Detecting pedestrians using patterns of motion and appearance.
\newblock {\em International journal of computer vision}, 63:153--161, 2005.

\bibitem{walambe2021multiscale}
Rahee Walambe, Aboli Marathe, and Ketan Kotecha.
\newblock Multiscale object detection from drone imagery using ensemble
  transfer learning.
\newblock {\em Drones}, 5(3):66, 2021.

\bibitem{a3}
Rahee Walambe, Aboli Marathe, Ketan Kotecha, George Ghinea, et~al.
\newblock Lightweight object detection ensemble framework for autonomous
  vehicles in challenging weather conditions.
\newblock {\em Computational Intelligence and Neuroscience}, 2021, 2021.

\bibitem{walambe2021lightweight}
Rahee Walambe, Aboli Marathe, Ketan Kotecha, George Ghinea, et~al.
\newblock Lightweight object detection ensemble framework for autonomous
  vehicles in challenging weather conditions.
\newblock {\em Computational Intelligence and Neuroscience}, 2021, 2021.

\bibitem{wilson2023argoverse}
Benjamin Wilson, William Qi, Tanmay Agarwal, John Lambert, Jagjeet Singh,
  Siddhesh Khandelwal, Bowen Pan, Ratnesh Kumar, Andrew Hartnett,
  Jhony~Kaesemodel Pontes, et~al.
\newblock Argoverse 2: Next generation datasets for self-driving perception and
  forecasting.
\newblock {\em arXiv preprint arXiv:2301.00493}, 2023.

\bibitem{wojek2009multi}
Christian Wojek, Stefan Walk, and Bernt Schiele.
\newblock Multi-cue onboard pedestrian detection.
\newblock In {\em 2009 IEEE Conference on Computer Vision and Pattern
  Recognition}, pages 794--801. IEEE, 2009.

\bibitem{yang2016exploit}
Fan Yang, Wongun Choi, and Yuanqing Lin.
\newblock Exploit all the layers: Fast and accurate cnn object detector with
  scale dependent pooling and cascaded rejection classifiers.
\newblock In {\em Proceedings of the IEEE conference on computer vision and
  pattern recognition}, pages 2129--2137, 2016.

\bibitem{zhang2015filtered}
Shanshan Zhang, Rodrigo Benenson, and Bernt Schiele.
\newblock Filtered feature channels for pedestrian detection.
\newblock In {\em Proceedings of the IEEE conference on computer vision and
  pattern recognition}, pages 1751--1760, 2015.

\end{thebibliography}
}

\appendix

\end{document}